\pdfoutput=1

\documentclass[11pt]{article}

\usepackage{ACL2023}

\usepackage{times}
\usepackage{latexsym}

\usepackage[T1]{fontenc}

\usepackage[utf8]{inputenc}

\usepackage{microtype}

\usepackage{inconsolata}

\usepackage{amssymb}

\usepackage{tikz}
\usetikzlibrary{positioning}
\usetikzlibrary{shapes}
\definecolor{DarkGreen}{rgb}{0.0, 0.5, 0.0}
\usepackage{booktabs}

\newcommand\blfootnote[1]{%
  \begingroup
  \renewcommand\thefootnote{}\footnote{#1}%
  \addtocounter{footnote}{-1}%
  \endgroup
}

\title{Learning to Transpile AMR into SPARQL}

  \author{ Mihaela Bornea$^{\clubsuit}$ , Ramon Fernandez Astudillo$^{\clubsuit}$, Tahira Naseem, Nandana	Mihindukulasooriya  \\ \textbf{Ibrahim	Abdelaziz}, \textbf{Pavan Kapanipathi},   \textbf{Radu Florian} \and \textbf{Salim Roukos} \\
        IBM Research AI \\ mabornea@us.ibm.com,   ramon.astudillo@ibm.com, tnassem@us.ibm.com \\ \{nandana.m, ibrahim.abdelaziz2\}@ibm.com \\ \{kapanipa, raduf, roukos\}@us.ibm.com} 
\begin{document}
\maketitle
\blfootnote{$\clubsuit$ These authors contributed equally to this work}

\begin{abstract}
We propose a transition-based system to transpile Abstract Meaning Representation (AMR) into SPARQL for Knowledge Base Question Answering (KBQA). This allows us to delegate part of the semantic representation to a strongly pre-trained semantic parser, while learning transpiling with small amount of paired data. We depart from recent work relating AMR and SPARQL constructs, but rather than applying a set of rules, we teach a BART model to selectively use these relations. Further, we avoid explicitly encoding AMR but rather encode the parser state in the attention mechanism of BART, following recent semantic parsing works. The resulting model is simple, provides supporting text for its decisions, and outperforms recent approaches in KBQA across two knowledge bases: DBPedia (LC-QuAD 1.0, QALD-9) and Wikidata (WebQSP, SWQ-WD).
\end{abstract}

\section{Introduction}

The task of Question Answering over Knowledge Bases (KBQA) \cite{Zou2014Natural,Vakulenko2019MessagePF,Diefenbach2020TowardsAQ} consists in finding the answer to natural questions by retrieving the information contained in a Knowledge Graph (KG). Compared to other automatic QA tasks, such as reading comprehension \cite{Rajpurkar2018KnowWY,naturalQuestions,clark-etal-2020-tydi}, 
KBQA has the advantage of producing answers backed by structured repositories of information, offering strong factual accuracy guarantees. 
Solving KBQA amounts to transforming a natural language question into some programming language, usually a query language such as SQL or SPARQL. It is thus considered a form of executable semantic parsing. We illustrate the KBQA task in Figure~\ref{sparql}.

\begin{figure}[!th]
{\footnotesize
\centering
\begin{center}
\begin{tikzpicture}[scale=0.75]

\node [right] (spa0) at (0,0.5) {\fontfamily{qcr}\selectfont PREFIX dbr:http://dbpedia.org/resource/};
\node [right] (spa0) at (0,0.0) {\fontfamily{qcr}\selectfont PREFIX dbo:http://dbpedia.org/ontology/};
\node [right] (spa0) at (0,-0.5) {\fontfamily{qcr}\selectfont PREFIX dbp:http://dbpedia.org/property/};
\node [right] (spa0) at (0,-1.0) {\fontfamily{qcr}\selectfont SELECT DISTINCT {\color{red} ?s} WHERE \{};
\node [right] (spa2) at (0.3,-1.5) {\fontfamily{qcr}\selectfont {\color{blue} ?i} <dbp:state> {\color{DarkGreen} <dbr:Maharashtra>}.};
\node [right] (spa3) at (0.3,-2.0) {\fontfamily{qcr}\selectfont {\color{blue} ?i} <dbo:sport> {\color{red} ?s}.};
\node [right] (spa4) at (0,-2.5) {\fontfamily{qcr}\selectfont\}};

\end{tikzpicture}
\end{center}
}
\caption{SPARQL graph for the question \textit{Name some sports played in institutions of Maharashtra?}. {\bf \color{red} ?s} is the {\bf \color{red} unknown variable} (related to \textit{sports}) and {\color{blue} ?i} is an {\color{blue} intermediate variable} (related to \textit{institution}), needed to relate the unknown with the {\color{DarkGreen} KG entity} (\textit{Maharashtra}). }
\label{sparql}
\end{figure}


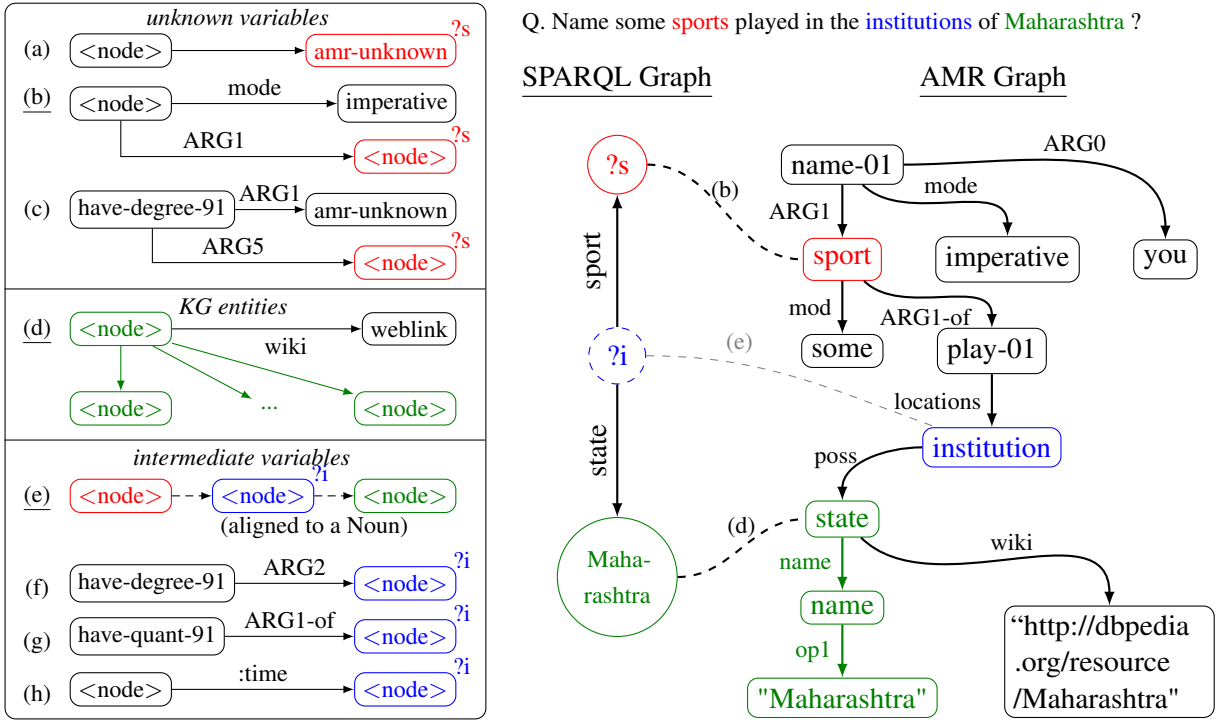
\begin{figure*}[!h]

\begin{tikzpicture}
\definecolor{DarkGreen}{rgb}{0.0, 0.5, 0.0}


\node [anchor=north west] (text) at (190pt,0) {\footnotesize Q. Name some {\color{red} sports} played in the {\color{blue} institutions} of {\color{DarkGreen}Maharashtra} ?};

\node [below = 30pt of text.north west,anchor=west] (q_label) {\underline{SPARQL Graph}};
\node [anchor=north][right = 70pt of q_label] (a_label){\underline{AMR Graph}};


\node [draw,circle,red, below = 10pt of q_label] (q_s) {?s};
\node [draw,dashed,circle, blue, below = 50pt of q_s] (q_i) {?i};
\node [draw, align=center,circle, DarkGreen, below = 50pt of q_i,text width= 1cm] (q_e) {\footnotesize Maha- rashtra};
\draw [-latex,thick] (q_i) -- node[above,rotate=90,pos=0.5]  { state} (q_e);
\draw [-latex,thick] (q_i) to[out=90,in=270] node[above,rotate=90,pos=0.5]  { sport} (q_s);

\node [draw,rounded corners,right = 50pt of q_s] (n) {name-01};
\node [draw,rounded corners,red,below = 20pt of n] (s) {sport};
\node [draw,rounded corners,right = 20pt of s] (i) {imperative};
\node [draw,rounded corners,right = 20pt of i] (y) {you};
\node [draw,rounded corners,below = 20pt of s] (s2) {some};
\node [draw,rounded corners,right = 20pt of s2] (p) {play-01};
\node [draw,rounded corners,blue,below = 20pt of p] (i2) {institution};
\node [draw,rounded corners,DarkGreen,below = 50pt of s2] (s1) {state};
\node [draw,rounded corners,DarkGreen,below = 20pt of s1] (n2) {name};
\node [draw,rounded corners,DarkGreen,below = 20pt of n2] (m) {"Maharashtra"};
\node [draw,rounded corners,right = 100pt of m.south,anchor=south,text width= 2.5cm] (w) {``http://dbpedia .org/resource /Maharashtra"};
\draw [-latex,thick] (n) -- node[left,pos=0.5]  {\footnotesize ARG1} (s);
\draw [-latex,thick] (n) to[out=-45,in=90] node[above,pos=0.5]  {\footnotesize mode} (i);
\draw [-latex,thick] (n) to[out=0,in=90] node[above,pos=0.5]  {\footnotesize ARG0} (y);
\draw [-latex,thick] (s) -- node[left,pos=0.5]  {\footnotesize mod} (s2);
\draw [-latex,thick] (s) to[out=-45,in=90] node[below,pos=0.4]  {\footnotesize ARG1-of} (p);
\draw [-latex,thick] (p) -- node[left,pos=0.5]  {\footnotesize locations} (i2);
\draw [-latex,thick] (i2) to[out=180,in=90] node[left,pos=0.5]  {\footnotesize poss} (s1);
\draw [-latex,thick,DarkGreen] (s1) -- node[left,pos=0.5]  {\footnotesize name} (n2);
\draw [-latex,thick] (s1) to[out=-45,in=90] node[above,pos=0.5]  {\footnotesize wiki} (w);
\draw [-latex,thick,DarkGreen] (n2) -- node[left,pos=0.5]  {\footnotesize op1} (m);

\draw [thick,dashed] (q_s) to[out=0,in=180] node[above,pos=0.5]  {\footnotesize (b)} (s);
\draw [dashed,gray] (q_i) to[out=0,in=160] node[above,pos=0.3]  {\footnotesize (e)} (i2);
\draw [thick,dashed] (q_e) to[out=0,in=180] node[above,pos=0.5]  {\footnotesize (d)} (s1);


{\footnotesize

\node [draw,rounded corners, minimum width =180pt, minimum height =270pt,anchor=north west] (box_unk) at (0,0) {};
\node [below right= -1pt and 50pt of box_unk.north west,anchor=north west] (label_unk)  {\textit{unknown variables}};

\draw [anchor=north west](0pt,-108pt) -- (180pt,-108pt);
\node [below right= 108pt and 62pt of box_unk.north west,anchor=north west] (label_ent)  {\textit{KG entities}};

\draw [anchor=north west](0pt,-165pt) -- (180pt,-165pt);
\node [below right= 165pt and 45pt of box_unk.north west,anchor=north west] (label_int)  {\textit{intermediate variables}};

\node [below right= 11pt and 4pt of box_unk.north west,anchor=north west] (rule_a)  {(a)};
\node [below = 18pt of rule_a.north] (rule_b)  {\underline{(b)}};
\node [below = 42pt of rule_b.north] (rule_c)  {(c)};
\node [below = 45pt of rule_c.north] (rule_d)  {\underline{(d)}};
\node [below = 63pt of rule_d.north] (rule_e)  {\underline{(e)}};
\node [below = 35pt of rule_e.north] (rule_f)  {(f)};
\node [below = 20pt of rule_f.north] (rule_g)  {(g)};
\node [below = 20pt of rule_g.north] (rule_h)  {(h)};
\node [draw,rounded corners,  right= 4pt of rule_a] (n_a)  {$<$node$>$};
\node [draw,rounded corners,red, right = 50pt of n_a] (unk_a) {amr-unknown};
\node [red,above right = 12pt of unk_a.north east,anchor = north east] (q_s1)  {\footnotesize ?s};
\draw [-latex] (n_a) -- node[left,pos=0.5]  {} (unk_a);

\node [draw,rounded corners, below = 20pt of n_a.west,anchor=west] (n_b) {$<$node$>$};
\node [draw,rounded corners, below = 20pt of unk_a.east,anchor=east] (unk_b) {imperative};
\node [draw,rounded corners, red, below = 20pt of unk_b.east,anchor=east] (n2_b) {$<$node$>$};
\node [red,above right = 12pt of n2_b.north east,anchor = north east] (q_s1)  {\footnotesize ?s};
\draw [-latex] (n_b) -- node[above,pos=0.5]  {\footnotesize mode} (unk_b);
\draw [-latex] (n_b) |- node[above,pos=0.7]  {\footnotesize ARG1} (n2_b);

\node [draw,rounded corners, below = 40pt of n_b.west,anchor=west] (d) {have-degree-91};
\node [draw,rounded corners, below = 20pt of n2_b.east,anchor=east] (unk_c) {amr-unknown};
\node [draw,rounded corners, red, below = 20pt of unk_c.east,anchor=east] (n_c) {$<$node$>$};
\node [red,above right = 12pt of n_c.north east,anchor = north east] (q_s1)  {\footnotesize ?s};
\draw [-latex] (d) -- node[above,pos=0.5]  {\footnotesize ARG1} (unk_c);
\draw [-latex] (d) |- node[above,pos=0.7]  {\footnotesize ARG5} (n_c);


\node [draw,rounded corners,DarkGreen, below = 45pt of d.west,anchor=west] (e) {$<$node$>$};
\node [draw,rounded corners, below = 25pt of n_c.east,anchor=east] (el) {weblink};
\node [draw,rounded corners,DarkGreen, below = 30pt of e.west,anchor=west] (e_n1) {$<$node$>$};
\node [DarkGreen,right = 30pt of e_n1] (dots) {...};
\node [draw,rounded corners,DarkGreen, below = 30pt of el.east,anchor=east] (e_n2) {$<$node$>$};
\draw [-latex] (e) -- node[below,pos=0.6]  {\footnotesize wiki} (el);
\draw [-latex,DarkGreen] (e) -- node[left,pos=0.5]  {} (e_n1);
\draw [-latex,DarkGreen] (e) -- node[left,pos=0.5]  {} (dots);
\draw [-latex,DarkGreen] (e) -- node[left,pos=0.5]  {} (e_n2);


\node [draw,rounded corners,red, below = 33pt of e_n1.west,anchor=west] (unk) {$<$node$>$};
\node [draw,rounded corners,blue, right = 15pt of unk] (ii) {$<$node$>$};
\node [blue,above right = 12pt of ii.north east,anchor = north east] (q_s1)  {\footnotesize ?i};

\node [below = 12pt of ii.west,anchor=west] (label) {\footnotesize (aligned to a Noun)};
\node [draw,rounded corners,DarkGreen, below = 33pt of e_n2.east,anchor=east] (entity) {$<$node$>$};
\draw [-latex,densely dashed] (ii) -- node[left,pos=0.5]  {} (entity);
\draw [-latex,densely dashed] (unk) -- node[left,pos=0.5]  {} (ii);
\node [draw,rounded corners, below = 33pt of unk.west,anchor=west] (d2) {have-degree-91};
\node [draw,rounded corners, below = 20pt of d2.west,anchor=west] (q) {have-quant-91};
\node [draw,rounded corners, below = 20pt of q.west,anchor=west] (t) {$<$node$>$};
\node [draw,rounded corners,blue, below = 33pt of entity.east,anchor=east] (ii2) {$<$node$>$};
\node [blue,above right = 12pt of ii2.north east,anchor = north east] (q_s2)  {\footnotesize ?i};
\node [draw,rounded corners,blue, below = 20pt of ii2.east,anchor=east] (ii3) {$<$node$>$};
\node [blue,above right = 12pt of ii3.north east,anchor = north east] (q_s3)  {\footnotesize ?i};
\node [draw,rounded corners,blue, below = 20pt of ii3.east,anchor=east] (ii4) {$<$node$>$};
\node [blue,above right = 12pt of ii4.north east,anchor = north east] (q_s4)  {\footnotesize ?i};
\draw [-latex] (d2) -- node[above,pos=0.5]  {\footnotesize ARG2} (ii2);
\draw [-latex] (q) -- node[above,pos=0.5]  {\footnotesize ARG1-of} (ii3);
\draw [-latex] (t) -- node[above,pos=0.5]  {:time} (ii4);
}
\end{tikzpicture}

\caption{
(Left) Rules used to align AMR with SPARQL. Matching AMR subgraph patterns (a-h) are assigned SPARQL constructs (variables, entities). For example rule (b) identifies AMR node `sport' as unknown variable because it is the object (\textsc{arg1}) of an `imperative' statement.  This is a significantly reduced subset of the rules in \cite{kapanipathi-etal-2021-leveraging}. Further, some of the rules (e-h for intermediate variables) are not deterministically applied but must be instead chosen by the model. 
(Right) Examples applying hard rules (b) and (d) and an optional rule (e). Dashed links represent alignments between AMR and SPARQL.}
\label{pathalgo}

\end{figure*}

While some large KBQA datasets exist~\cite{Yu2018SpiderAL}, the amount of paired examples, i.e. aligned natural language and query language pairs, is generally scarce \cite{Trivedi,Usbeck}. Furthermore, human language exhibits large variability, and obtaining enough training pairs for specific domains or infrequent natural language formulations requires large annotation investments. Often, real world implementations of KBQA end up containing a number of domain specific hand-crafted rules that can be costly to maintain and expand. The need to manipulate a formal representation with only a few examples, makes this task harder to learn compared to other QA tasks.

In order to mitigate the data availability problem \cite{kapanipathi-etal-2021-leveraging} proposed to delegate part of the task to an Abstract Meaning Representation (AMR) parse that incorporates additional semantic information. This work identifies associations between certain AMR nodes and SPARQL entities and variables. Unfortunately, these AMR to SPARQL mappings are imperfect, suffering from low coverage and granularity mismatch. 
Another concern with AMR-based approaches is the encoding of the AMR itself which implies an additional learning burden.

To address these limitations, in this work we propose a new approach to leverage AMR parsing for KBQA that \textit{learns to transpile} AMR into the SPARQL query language. The contributions can be summarized as follows

\begin{itemize}

\item We develop a state machine and an oracle that transpiles AMR into SPARQL and we learn to imitate this oracle with BART~\cite{lewis2019bart} for KBQA tasks.

\item This oracle leverages known relations (i.e. similarities) between AMR and SPARQL \cite{kapanipathi-etal-2021-leveraging}, but rather than applying them deterministically as in prior work, we teach the model when to use them.

\item We show that it is not necessary to encode AMR directly, but rather encoding the transpiler state through attention masking as in \cite{emnlp2020stacktransformer} suffices.

\item The resulting transpiler improves upon recent KBQA approaches across Dbpedia and Wikidata. It outperforms \cite{kapanipathi-etal-2021-leveraging}, by $10$ points on LcQuAD 1.0 and matches it on QALD-9, while being simpler and exploiting similar inductive biases. It also outperforms \cite{sygma}, by $2$ points on WebQSP-WD and $4$ points on SWQ-WD.

\end{itemize}

\section{AMR to SPARQL Machine and Oracle}

Here we show how transition-based parsing can be used to traspile AMR to SPARQL.

\subsection{A Transition-based Transpiler}

\begin{figure*}[!h]
{\footnotesize
\centering
\begin{center}
\begin{tikzpicture}[scale=0.7]
\def \tokx {-1.9}
\def \tokax {-0.4} 
\def \tokbx {0.75} 
\def \tokcx {1.95} 
\def \tokdx {3.25} 
\def \tokex {4.20} 
\def \tokfx {5.60} 
\def \tokgx {6.95} 
\def \tokhx {8.45} 
\def \tokix {9.8} 

\draw(\tokx ,0.00) node {$w$:};
\draw(\tokax,0.00) node {\strut Name};
\draw(\tokbx,0.00) node {\strut some};
\draw(\tokcx,0.00) node (sports) {\strut sports};
\draw(\tokdx,0.00) node (played) {\strut played};
\draw(\tokex,0.00) node {\strut in};
\draw(\tokfx,0.00) node (institutions) {\strut institutions};
\draw(\tokgx,0.00) node {\strut of};
\draw(\tokhx,0.00) node (Maharashtra) {\strut Maharashtra};
\draw(\tokix,0.00) node {?};
\draw(\tokx,6.30) node {$g$:};
\node [draw,rounded corners] (n) at (\tokax,6.30) {n/name-01};
\node [draw,rounded corners] (i1) at (\tokex,4.80) {imperative};
\node [thick,draw=red,rounded corners] (s) at (\tokcx,4.80) {{\color{red} s/sport}};
\node [draw,rounded corners] (y) at (\tokgx,4.80) {y/you};
\node [draw,rounded corners] (s2) at (\tokbx,3.30) {s2/some};
\node [draw,rounded corners] (p) at (\tokdx,3.30) {p/play-01};
\node [thick,draw=blue,rounded corners] (i) at (\tokfx,3.30) {{\color{blue} i/institution}};
\node [thick,draw=DarkGreen,rounded corners] (s1) at (\tokhx,3.30) {{\color{DarkGreen} s1/state}};
\node [thick,draw=DarkGreen,rounded corners,anchor=west] (el0) at (11.00,0.50) {{\color{DarkGreen} "http://dbpedia.org/resource/Maharashtra"}};
\node [thick,draw=DarkGreen,rounded corners] (n1) at (\tokhx,2.20) {{\color{DarkGreen} name}};
\node [thick,draw=DarkGreen,rounded corners] (0) at (\tokhx,1.10) {{\color{DarkGreen} "Maharashtra"}};
\draw [-latex,thick,right,right] (n) to[out=0,in=160] node[above,pos=0.6] {\fontsize{6}{6}\selectfont ARG0} (y);
\draw [-latex,thick,right,right] (n) -- node[left,pos=0.5] {\fontsize{6}{6}\selectfont ARG1} (s);
\draw [-latex,thick,right,right] (s) -- node {\fontsize{6}{6}\selectfont ARG1-of} (p);
\draw[-latex, thick, DarkGreen] (s1) to[out=0,in=150] node[midway,right] {\footnotesize wiki} (11.0,0.50);

\draw[-latex, thick] (p) to[out=330,in=210] node[midway,below] {\footnotesize location} (i);
\draw[-latex, thick] (i) to[out=330,in=210] node[midway,below] {\footnotesize poss} (s1);

\draw [-latex,thick,right,DarkGreen,right] (s1) -- node {\footnotesize name} (n1);
\draw [-latex,thick,right,right] (s) -- node [left,pos=0.4] {\footnotesize mod} (s2);
\draw [-latex,thick,right,right] (n) -- node[above,pos=0.8] {\fontsize{8}{8}\selectfont mode} (i1);
\draw [-latex,thick,right,DarkGreen,right] (n1) -- node {\footnotesize op1} (0);

\draw [densely dashed,red] (sports) -- (s);
\draw [densely dashed] (played) -- (p);
\draw [densely dashed,blue] (institutions) -- (i);
\draw [densely dashed,DarkGreen] (Maharashtra) -- (0);


\def \bartx {13.7}
\def \barty {2.0}
\def \bartyb {1.5}

\draw[draw=black,dashed,rounded corners, anchor=west] (\bartx-1.9,\barty -0.4) rectangle ++(9,4.9);

\draw({\bartx + 2.6},{\bartyb + 4.6}) node[anchor=west] (inputsentence) {\scriptsize locationCity};
\draw({\bartx + 2.6},{\bartyb + 4.2}) node[anchor=west] (inputsentence) {\scriptsize homeTown};
\draw({\bartx + 2.6},{\bartyb + 3.8}) node[anchor=west] (inputsentence) {\scriptsize \underline{state}};
\draw({\bartx + 2.6},{\bartyb + 3.4}) node[anchor=west] (inputsentence) {\scriptsize populationTotal};
\draw({\bartx + 2.6},{\bartyb + 3.0}) node[anchor=west] (inputsentence) {\scriptsize birthPlace};
\draw [-] ({\bartx + 2.6},{\bartyb + 2.9}) -- ({\bartx + 2.6},{\bartyb + 4.8});
\draw [-] ({\bartx + 2.6},{\bartyb + 3.8}) -- ({\bartx + 1.5},{\bartyb + 4.5});


\node[rectangle,draw, minimum width=2mm, minimum height=2mm, fill=red!20] (r) at ({\bartx - 1.0},{\barty + 1.3}) {};
\node[rectangle,draw, minimum width=2mm, minimum height=2mm, fill=red!20] (r) at ({\bartx - 0.2},{\barty + 1.3}) {};
\node[rectangle,draw, minimum width=2mm, minimum height=2mm, fill=red!20] (r) at ({\bartx + 0.7},{\barty + 1.3}) {};
\node[rectangle,draw, minimum width=2mm, minimum height=2mm, fill=red!20] (r) at ({\bartx + 1.6},{\barty + 1.3}) {};

\node[rectangle,draw, minimum width=2mm, minimum height=2mm, fill=red!20] (r) at ({\bartx + 2.3},{\barty + 1.3}) {};
\node[rectangle,draw, minimum width=2mm, minimum height=2mm, fill=blue!10] (aa1) at ({\bartx + 3.2},{\barty + 1.3}) {};
\node[rectangle,draw, minimum width=2mm, minimum height=2mm, fill=red!20] (r) at ({\bartx + 4.3},{\barty + 1.3}) {};
\node[rectangle,draw, minimum width=2mm, minimum height=2mm, fill=blue!10] (aa2) at ({\bartx + 5.4},{\barty + 1.3}) {};

\node[rectangle,draw, minimum width=2mm, minimum height=2mm, fill=red!20] (n3) at ({\bartx + 1.6},{\barty + 3.1}) {};

\node[rectangle,draw, minimum width=2mm, minimum height=2mm, fill=blue!10] (n2) at ({\bartx + 1.6},{\barty + 2.6}) {};
\draw [-] (aa1) -- (n2);
\draw [-] (aa2) -- (n2);

\node [draw,circle,DarkGreen,text width=0.6cm] (ma) at ({\bartx + 6},{\bartyb + 3.8}) {{\color{DarkGreen} \tiny Maha- rashtra}};
\draw [-,dashed] ({\bartx + 6},{\bartyb + .6}) -- (ma);

\draw(\bartx - 0.2,{\barty + 4}) node (inputsentence) {\scriptsize SELECT REDUCE};
\node[rectangle,draw, minimum width=2.5cm, minimum height=0.5cm, anchor=west, fill=blue!10] (r) at ({\bartx - 1.7},{\barty + 3.3}) {BART decoder};
\node[rectangle,draw, minimum width=6cm, minimum height=0.5cm, anchor=west, fill=blue!10] (r) at ({\bartx - 1.7},{\barty + 0.7}) {BART encoder};
\draw({\bartx + 2.5},\barty) node (inputsentence) {\scriptsize Name some sports played in {\color{blue} institutions} of {\color{DarkGreen} Maharashtra} ?};


\def \pxa {-2.0}
\def \pxb {2.3}
\def \pxc {9.7}
\def \pxd {11.7}

\draw[draw=black,dashed,rounded corners] (\pxa-0.2,-0.5) rectangle ++(23,-6);

\node [right] (title0) at (\pxa,-1.0) {\strut \textbf{Path Stack:}};
\node [right] (title1) at (\pxb,-1.0) {\strut \textbf{Supporting Text:}};
\node [right] (title2) at (\pxc,-1.0) {$a$: \strut \textbf{Actions:}};
\node [right] (title2) at (\pxd+1,-1.0) {$s$: \strut \textbf{SPARQL:}};

\node [right,draw] (path1) at (\pxa,-3.0) {\strut {\color{DarkGreen} s1} {\color{blue} i} p {\color{red} s}};
\node [right,draw, fill=black!10] (path2) at ({\pxa+1.7},-3.0) {\strut s1 i};
\node [right,draw, fill=black!10] (path3) at ({\pxa+2.8},-3.0) {\strut i p s};

\node [right,draw] (path2) at ({\pxa+1.7},-4.0) {\strut {\color{DarkGreen} s1} {\color{blue} i}};
\node [right,draw, fill=black!10] (path3) at ({\pxa+2.8},-4.0) {\strut i p s};

\node [right,draw] (path3) at ({\pxa+2.8},-5.0) {\strut {\color{blue} i} p {\color{red} s}  };


\node [right] (ratio0) at (\pxb,-2.0) {};
\node [right] (ratio1) at (\pxb,-3.0) {sports played institutions Maharashtra};
\node [right] (ratio2) at (\pxb,-4.0) {institutions Maharashtra};
\node [right] (ratio3) at (\pxb,-5.0) {sports played institutions};
\node [right] (act0) at (\pxc,-2.0) {SELECT};
\node [right] (act1) at (\pxc,-3.0) {REDUCE};
\node [right] (act2) at (\pxc,-4.0) {state};
\node [right] (act3) at (\pxc,-5.0) {sport};
\node [right] (act4) at (\pxc,-6.0) {CLOSE};

\node [right] (spa0) at (\pxd,-2.0) {\fontfamily{qcr}\selectfont SELECT DISTINCT {\color{red} ?s} WHERE \{};
\node [right] (spa1) at (\pxd,-3.0) {};
\node [right] (spa2) at (\pxd,-4.0) {\fontfamily{qcr}\selectfont {\color{blue} ?i} <dbp:state> {\color{DarkGreen} <dbr:Maharashtra>}.};
\node [right] (spa3) at (\pxd,-5.0) {\fontfamily{qcr}\selectfont {\color{blue} ?i} <dbo:sport> {\color{red} ?s}.};
\node [right] (spa4) at (\pxd,-6.0) {\fontfamily{qcr}\selectfont\}};


\end{tikzpicture}
\end{center}
}
\caption{Top-left: LC-QuAD train sentence $w$: \textit{Name some sports played in institutions of Maharashtra?} aligned to its AMR graph $g$ (system input) and $3$ relevant subgraphs identified by applying Figure~\ref{pathalgo} (c, d, e): {\bf \color{red} unknown variable} (\textrm{imperative} root), {\bf \color{blue} (optional) secondary variable} (entity-adjacent nominal), {\bf \color{DarkGreen} linked entity} (\textrm{wiki}). Top-right: time-step 3 of decoding of same sentence indicating AMR/KG-based constrained decoding and attention masking. Bottom box: Full oracle for same sentence including implicit machine state defined by the AMR path stack, explicit machine state defined by the supporting text, oracle action sequence $a$ and resulting SPARQL $s$ (system output). 
}
\label{figure1b}
\end{figure*}
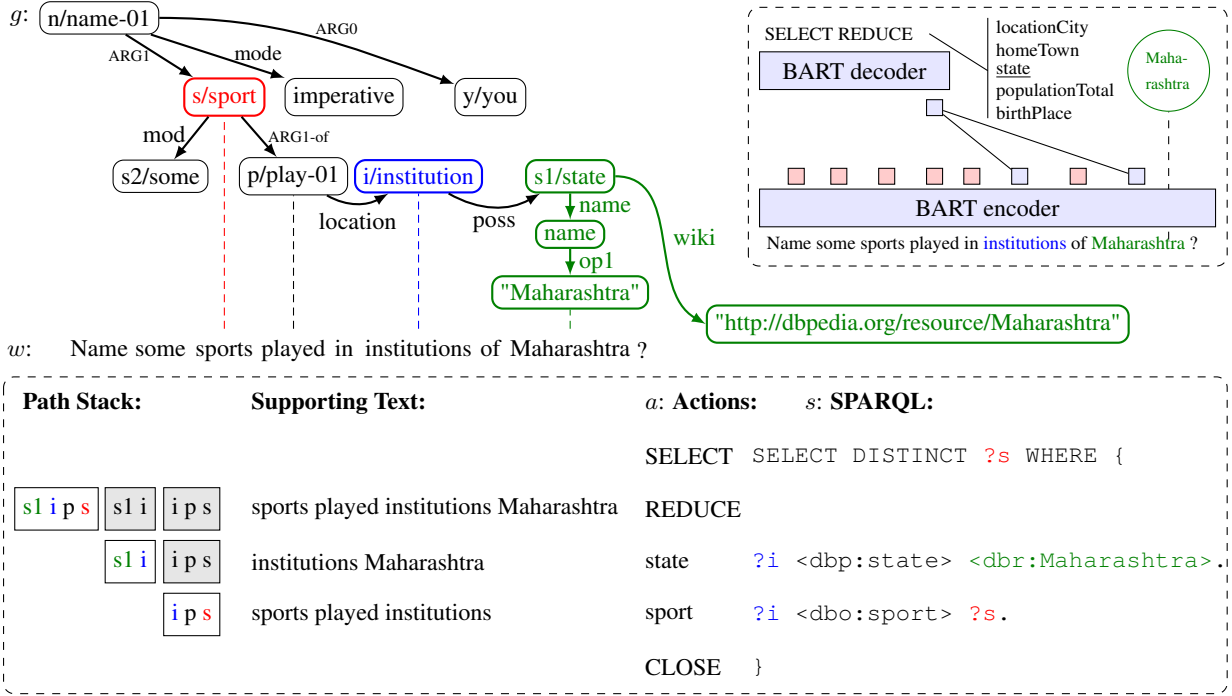

The objective is to learn to predict the SPARQL query $s$ corresponding to the natural language question $w$, by transforming its AMR parse $g$. 

At its core, a transition-based transpiler learns to predict a sequence of actions $a$ that applied to a parameter-less state machine acting on $g$ yields $s$.  
\begin{equation}
s = M(a, g). 
\end{equation}
The sequence of actions is given by a rule-based oracle that, given the original question $w$, its AMR $g$ and the gold SPARQL $s$, yields the action sequence. 
\begin{equation}
a = O(s, g, w).
\label{eq:oracleactions}
\end{equation}
The oracle, detailed in Section~\ref{sec:oracle}, requires alignments between the AMR $g$ and SPARQL $s$ graphs but it is only needed to generate training samples $(w, g, a) \sim \mathcal{D}$, through following procedure:

\begin{itemize}
\item draw sentence and SPARQL train pair $(w, s)$ from a KBQA training corpus
\item parse $w$ into its AMR graph $g$\footnotemark\footnotetext{Throughout this work we always use the APT parser \cite{zhou-etal-2021-amr} to obtain AMR graphs.} 
\item obtain oracle sequence $a = O(s, g, w)$
\end{itemize}

With these, one can use the oracle as a teacher to train a sequence to sequence model, e.g. BART~\cite{lewis-etal-2020-bart}, with a conventional cross entropy loss
\begin{equation}
\hat{\theta} = \arg\max_{\theta}\{\mathbb{E}_{(w, g, a) \sim \mathcal{D}}\{\log p(a \mid w, g; \theta)\}\}.
\label{eq:machine}
\end{equation}
At test time, we predict the SPARQL as $\hat{s} = M(\hat{a}, g)$, where $g$ is obtained by parsing $w$ and $\hat{a}$ is obtained with conventional decoding.
\begin{equation}
\hat{a} = \arg\max_{a}\{p(a \mid w, g; \theta)\}
\end{equation}

It is important to note that the use of a state machine generates a strong inductive bias, imposing some specific way in which the sequence to sequence problem can be solved. In this case we leverage the bias to make the transpiler aware of AMR path information, but it comes at the penalty of not fully being able to recover all SPARQL queries i.e. for some queries. 
\begin{equation}
s \neq M(O(s, g, w), g)
\end{equation}
\noindent This is analyzed in Section~\ref{oracleperformance}.

\subsection{KBQA Oracle and Machine}
\label{sec:oracle}

This section describes how to formulate the state machine $s=M(a, g)$ and oracle $a=O(s, g, w)$. The basic idea is that given alignments between AMR and SPARQL graphs, as depicted in Figure~\ref{pathalgo} (right), it is possible to sequentially proceed over a subset of paths in the AMR graph and transform them into components of the SPARQL query, as shown in Figure~\ref{figure1b} (bottom).

Initially, all AMR paths to be processed are put in a stack, Figure~\ref{figure1b} (bottom left). At each time step one of the following actions is chosen by the oracle

\begin{itemize}
\item \textrm{\{\textrm{SELECT}, \textrm{ASK}, \textrm{COUNT}\}}: Generate the query header from a closed vocabulary
\item \textrm{<KG relation>}: Produce the KG relation for path at the top of the stack and \textrm{REDUCE}
\item \textrm{REDUCE}: Pop path at the top of the stack without predicting any KG relation.
\item \textrm{\{\textrm{CLOSE}\}}: Close the machine
\end{itemize}

All actions are special symbols added to BART's vocabulary with the exception of KG relations, which allows free text generation. The stack initially contains all AMR paths between entities in the KG ({\color{DarkGreen} green}) and the unknown variable in the query ({\color{red} red}) and optional paths involving secondary variables ({\color{blue} blue}), if there are any in the AMR. As long as all the AMR subgraphs matching SPARQL components are present in $g$, the transpiling oracle is able to generate SPARQL queries with any number of entities and with the presence of secondary variables i.e. multi-hop questions\footnotemark\footnotetext{Section~\ref{oracleperformance} for an empirical analysis of oracle coverage in KBQA datasets.}. The \textrm{REDUCE} action allows the oracle to ignore a given AMR path, which in practice controls the application of rules (e-h). This is useful to compensate for false positive cases where this rule generates non-existing secondary unknowns.

As an example: In Figure~\ref{figure1b}, the path at the top of the stack is [{\fontfamily{qcr}\selectfont s1 i p s}]. This path is determined by the alignment of the unknown variable {\fontfamily{qcr}\selectfont s} in the AMR to the SPARQL projected variable {\fontfamily{qcr}\selectfont ?s} and the alignment of the subgraph {\fontfamily{qcr}\selectfont s1} to the entity {\fontfamily{qcr}\selectfont Maharashtra} (rules b, d). Since there is no gold SPARQL triple connecting {\fontfamily{qcr}\selectfont ?s} and {\fontfamily{qcr}\selectfont Maharashtra}, the action for this path is \textrm{REDUCE}. For the second AMR path [{\fontfamily{qcr}\selectfont s1 i}], there is a gold SPARQL triple matching the entity {\fontfamily{qcr}\selectfont Maharashtra} and an intermediate SPARQL variable {\fontfamily{qcr}\selectfont ?i} (obtained by rule e) thus the action for this path is the KG relation {\fontfamily{qcr}\selectfont state}.  

One fundamental advantage of the proposed approach, is that by construction it aligns sentence $w$, AMR $g$ and SPARQL $s$. This information can be used during decoding to restrict the output vocabulary of $p(a \mid w)$. For example, we can enforce header and closing operations only on a full and empty stack respectively. Further, we can also query the KG with nodes involved in the path on the top of the stack, to restrict the possible relations to predict. For example in Figure~\ref{figure1b}, bottom, to predict the KG relation {\fontfamily{qcr}\selectfont state} we restrict it to incoming or outgoing KG relations of the node {\fontfamily{qcr}\selectfont s1, Maharashtra}. Finally, we also obtain textual cues to predict the KG relation, see \textbf{Supporting Text} in Figure~\ref{figure1b}. In Section~\ref{state} we describe how textual cues are incorporated into the model, allowing to avoid encoding AMR explicitly.

\section{BART Transpiler with Inductive Bias}
\label{state}

We parametrize $p(a \mid w, g ; \theta)$ with a modified BART model \cite{lewis2019bart}, but the approach is applicable to sequence-to-sequence Transformers \cite{vaswani2017attention}. 
We dynamically mask cross-attention to encode the transpiler state, described in Section~\ref{sec:oracle}.
We also dynamically masking of the output vocabulary to force the decoder to generate only the valid actions for every state.

\subsection{Encoding AMR through Transpiler State}
\label{crossattention}

In order to encode both sentence $w$ and AMR $g$ in $p(a \mid w, g; \theta)$ one simple option could be to concatenate both $w$ and $g$ as inputs to BART. Given the quadratic cost of Transformer's attention, the large size of AMRs and the redundancies between $w$ and $g$ this is likely too expensive. 

Our model does not need to encode the entire AMR. It suffices to encode the parser state in the cross-attention heads of the Transformer. Unlike prior work~\cite{emnlp2020stacktransformer,zhou-etal-2021-amr,zhou-etal-2021-structure}, our transpiler stack does not contain the question words directly but contains AMR paths  which we use to represent the transpiler state. A key element is the alignment between the AMR nodes and words in the question, which allows us to identify the supporting text for every path. 

For example, in Figure~\ref{figure1b}, top-right we show the BART decoder at the third decoding step where the path at the top of the stack is [{\fontfamily{qcr}\selectfont s1 i}]. The supporting text for this path is $institutions~Maharashtra$. 

The encoder takes as input the question and we dedicate some heads to attend to the words in the supporting text for the path at the top of the stack.

 These structured head weights are given by 

\begin{equation}
W^{\mbox{\scriptsize cross-att}}_{it} \propto e^{\frac{1}{d} (K \cdot h^w_i)^T \cdot (Q \cdot h^{a}_t) + m_2(a_{<t}, w, g)_{i}}
\end{equation}
where $h^{w}_i$ and $h^{a}_t$ are encoder and decoder representations for $i^{th}$ word and $t^{th}$ prediction, and $Q$, $K$, $d$ are query, value and scaling weights. $m_2()_{i}$ masks $h^{w}_i$ unless it is aligned to the path of $g$ for which we are going to predict relation at step $t$.

Using the additional supporting text for every transpiler decision, helps the explainability of the model decisions, but also plays an important role in performance, as shown in Section~\ref{oracleperformance}.

\subsection{Constrained Decoding}
\label{constraineddecoding}

We use the state machine described in Section~\ref{sec:oracle} to forbid certain actions at each time step $t$ through
\begin{equation}
p(a_t \mid a_{<t}, w, g ;\theta) \propto e^{f(a_{<t}, w, \theta) + m(a_{<t}, w, g)}
\end{equation}
where $f(a_{<t}, w)$ is the output layer of BART without the softmax and $m(a_{<t}, w, g)$ is a state-machine dependent mask that is set to $-\infty$ to forbid actions. $m()$ is a deterministic function of the input sentence $w$, its AMR $g$ and action history $a_{<t}$ until step $t-1$. 

For the proposed model, masking is used to perform header actions only at the beginning of the action sequence and closing on an empty stack. 
In addition, when the AMR path at the top of the stack contains entity nodes, the KG is queried and the mask is set to restrict the actions to the appropriate KG relations. We show such an example in the BART decoder in Figure~\ref{figure1b}, top-right, where the actions are restricted to the KG relations for the entity $Maharashtra$. The relation's prefix\footnote{PREFIX dbo: http://dbpedia.org/ontology/, PREFIX dbp: http://dbpedia.org/property/ } and direction is also obtained in this process. 

These constrains are applied at decoding time only. As shown in the Section~\ref{experimentalsetup}, constrained decoding has a fundamental effect on performance.

\section{Experimental Setup}
\label{experimentalsetup}


\subsection{Datasets}
We evaluated our system on two knowledge graphs: DBpedia and Wikidata. For DBpedia we used the LC-QuAD 1.0 and QALD-9 datasets following the partitions in \cite{kapanipathi-etal-2021-leveraging}.
For Wikidata we used the SWQ-WD and WebQSP-WD datasets with splits from \cite{sygma}. As in prior work, we report the Macro F1 score comparing the gold answers to the answers that TransQA generates when executing its predicted SPARQL. We compute the F1 score for every question and we report the average over all questions.

\paragraph{LC-QuAD 1.0} \cite{Trivedi} contains 4,000 questions for training and 1,000 questions for test, created from templates. We tune the hyper parameters of our model on a random sample of 200 questions from the training set. 
LC-QuAD 1.0 predominantly focuses on multi-relational questions, aggregation (e.g. COUNT) and simple questions. Although Lc-QuAD has quantitative questions, the KG contains relations that can answer these queries without the need for specific SPARQL constructs e.g. {\fontfamily{qcr}\selectfont largestCity}.

\paragraph{QALD-9} \cite{Usbeck} has 408 training and 150 test questions concerning DBpedia.  We created a randomly chosen set of 98 questions as development set. QALD-9 contains more complex queries than LC-QuAD 1.0 including quantitative questions, time constructs and other.

\paragraph{SWQ-WD} \cite{swqwd} is a KBQA dataset adapted to Wikidata with 14,894 train and 5,622 test examples. We held out 200 examples from the train set for development. This dataset contains simple queries, with only one relation.

\paragraph{WebQSP-WD} \cite{webqspwd} has 2,880 train and 1,033 test set questions based on Wikidata. It includes more complex queries, often involving more than one relation.

\subsection{TransQA Model and Training Setup}

The model, henceforth referred to as TransQA, uses a modified BART \cite{lewis-etal-2020-bart} implemented in \cite{wolf-etal-2020-transformers} to parameterize $f(a_{<t}, w)$ and trained to learn to imitate the oracle. We implement constrained decoding at test time and parser state encoding by masking $8/16$ cross-attention heads both at train and test time, see Section \ref{state} for details.

For LC-QuAD 1.0 the model was trained with the LC-QuAD 1.0 train questions for 13 epochs and using a learning rate of $5\times10^{-5}$. 
The QALD-9 model is trained with both   LC-QuAD 1.0 and QALD-9 training examples. We upsample the QALD-9 5 times to balance between the two datasets and we trained the model for 16 epochs using a learning rate of $4\times10^{-5}$. 
For SWQ-WD we used its train set and trained the system for 9 epochs with learning rate $5\times10^{-5}$.
WebQSP-WD does not provide SPARQL queries and we use a combination of LC-QuAD 2.0 \cite{lcquad2} and SWQ-WD to train the system. LC-QuAD 2.0 \cite{lcquad2} is a template-based dataset based on Wikidata. It has some known quality issues \cite{diomedi2022entity} and  does not have a baseline to compare against. Therefore, we used its training data only to build our models. 
From LC-QuAD 2.0, we use only training examples with direct relations.  The final train set contains 12,900 LC-QuAD 2.0 and 14,693 SWQ-WD examples. We trained the system for 5 epochs with learning rate $5\times10^{-5}$. 

For all models we set the max input sequence length to $64$ tokens, max target sequence length to $32$ tokens and beam size to $4$. We used the Adam optimizer with standard parameters and trained on a V100 Nvidia GPU. All hyper-parameters were determined based on the dev set F1 score. The number of epochs and learning rate we determined from ranges $[3,18]$ and $[10^{-5},10^{-4}]$ respectively.

\subsection{AMR Parsing and Entity Linking}

We used the Action Pointer Transformer (APT) transition-based AMR parser \cite{zhou-etal-2021-amr} which provides alignments between AMR graph nodes and surface tokens. The model was trained combining AMR3.0 with the QALD-9-AMR train dataset \cite{lee2021maximum} to adapt it to the question domain\footnotemark\footnotetext{AMR train has no overlap with QALD-9 dev or test.}. Although AMR graphs can include entity linking, BLINK \cite{Wu2020ScalableZE} was also applied as a post-processing stage to AMR parsing. 
We run BLINK separately and then attach the {\fontfamily{qcr}\selectfont :wiki} edge to the most suitable node in the AMR.
 Linking the AMR nodes to KG entities is attained by matching the span of text aligned to a subgraph with the mention via a greedy set of checks. Attachment to conventional named entities is attempted first. Edit distance and fuzzy match search are also used to find a suitable alignment.

The rule (e) in Figure~\ref{pathalgo} requires Part of Speech tagging for which we use Spacy. 

See Section~\ref{sec:oracle} for information on how alignments are used to determine parser state. The same mechanism is used to produce the oracle action sequences for training.

\section{Comparison with Related Work}
\label{related}

\begin{table*}[!ht]
    \centering
    \begin{tabular}{l||c|c|c||c|c|c|c}
    \toprule
     &
     \multicolumn{1}{c|}{EL} &
     \multicolumn{1}{c|}{RL} &
     \multicolumn{1}{c||}{SI} &
      \multicolumn{1}{c|}{LCQ1} &
      \multicolumn{1}{c|}{Q9} & 
      \multicolumn{1}{c|}{WQSP} &
      \multicolumn{1}{c}{SWQ}\\
	\midrule
	NSQA \cite{kapanipathi-etal-2021-leveraging} &BLINK&BERT& rules & 44.5\footnotemark & 30.9 & - &-\\
	EDGQA \cite{edgqa} &Ensemb.&BERT& rules & 53.1 & \bf{32}.0 & -&-\\
	STaG-QA \citep{ravishankar2021twostage} &BLINK&-& BERT & 51.4 &- & - &*\footnotemark\\
	SYGMA \cite{neelam2021sygma}  &BLINK&BERT& rules & 47.0 & 29.0 & 31.0 &  44.0\\
	\hline
	Seq2Seq &BLINK&-& BART & 20.5 & 22.6 & 30.3 & \bf{50.3} \\
	TransQA &BLINK&-& BART & \bf{54.5} & 31.4 & \bf{33.0} & 48.1 \\
	\bottomrule
\end{tabular}
    \caption{TransQA comparison with other recent KBQA approaches. Including Entity Linking (EL), Relation Linking (RL) and the SPARQL inference (SI) components. TransQA uses a single module for relation linking and SPARQL inference. Macro F1 scores for SPARQL results on the test set to LC-QuAD1.0 (LCQ1), QALD-9 (Q9), WebQSP-WD (WQSP), SWQ-WD (SWQ)
    } 
    \label{tab:kbqa-results}
\end{table*}

\begin{table*}[h]
    \centering
    \begin{tabular}{c|c||cc|cc}
    \toprule
     &&
      \multicolumn{2}{c|}{LC-QuAD 1.0} &
      \multicolumn{2}{c}{QALD-9} \\
     
    AMR parsing$^\dagger$ &Entity Linking &    Oracle (F1)   &  TransQA  (F1)    &   Oracle  (F1)  & TransQA  (F1)  \\
	\midrule
    APT &BLINK & 68.7  & 47.4 & 67.0 & 53.6\\
    %
    APT &Gold  & 78.7  & 63.1 & 77.5 & 58.6\\
	\bottomrule
\end{tabular}
    \caption{Oracle and TransQA performances as well as effect of gold Entity Linking, measured by macro F1 of SPARQL results on the dev set of LC-QuAD 1.0 and QALD-9. Oracle performance with gold Entity Linking can be seen as reflecting transpiler coverage i.e. how much of the SPARLQ patterns are captured and an approximate upper bound for TransQA performance. $^\dagger$Note that entity Linking is part of AMR but is separated here for clarity.}
    \label{tab:oracle-results}
\end{table*}

\begin{table*}[t]
    \centering
    \begin{tabular}{l||c|c}
    \toprule
     &
      \multicolumn{1}{|c|}{LC-QuAD 1.0} &
      \multicolumn{1}{c}{QALD-9} \\
	\midrule
	TransQA~~~mask $8$ heads &  \textbf{56.5}  & \textbf{53.6} \\
	\midrule
	TransQA~~~no mask & 52.1 &  49.5 \\
	TransQA~~~mask 5 heads & 56.0 &	50.5 \\
	TransQA~~~mask 12 heads & 55.8 & 52.8 \\
	\midrule
	TransQA~~~no constrained decoding using KG & 46.3 &  41.8 \\
	\bottomrule
\end{tabular}
    \caption{Ablation study measuring the effect of constrained decoding and parser state encoding on BART. Measured by the macro F1 of SPARQL results on the dev set.}
    \label{tab:ablation-results}
\end{table*}

We compare TransQA against a number of recent neural approaches for KBQA trained in comparable conditions. We excluded approaches using gold entity linking and relation linking \cite{banerjee2022modern} since, both tasks have a strong impact in performance. We also include a BART baseline to control for the effect of pre-trained models since most other recent methods use BERT~\cite{Devlin2019BERTPO}. 

\textbf{NSQA} \cite{kapanipathi-etal-2021-leveraging} Uses a pipeline of AMR parsing (StackTransformer \cite{emnlp2020stacktransformer}), entity linking (BLINK \cite{Wu2020ScalableZE}), rule-based transpiling of AMR, BERT based Relation linking (SemRel~\cite{semrel}) and a final integrator module producing SPARQL from competing hypothesis. TransQA, borrows a significantly reduced amount of rules from this work for alignment, but replaces this pipeline by a simpler learnt transpilation of AMR using a modified BART \cite{lewis-etal-2020-bart}.

\textbf{SYGMA} \cite{sygma} is another multi-component pipeline with final integrator and using the same entity and relation linking as NSQA. It leverages lambda calculus as intermediate representation and features improved handling of temporal relations and an extended experimental setup.

\textbf{EDGQA} \cite{edgqa} proposes a custom Entity Description Graph (EDG) to represent the structure of complex questions, rather than relying on established formalisms such as AMR. It applies multiple decomposition rules and BERT-based re-scoring to produce SPARQL. It also uses an ensemble of three entity linking tools: Dexter \cite{Ceccarelli2014Dexter2}, EARL \cite{Dubey2018EARL} and Falcon \cite{sakor-etal-2019-old} as one entity retriever. EDGQA also uses a relation linker trained with LC-QuAD 1.0 and QALD-9.

\textbf{STaG-QA} \citep{ravishankar2021twostage} is a two stage text to SPARQL system that first generates a query skeleton learned end-to-end with
a BERT model and then performs beam search to find optimal grounding of the skeleton into a target KG. The paper also provides results for silver data training which are left-out for comparability purposes.

\textbf{seq2seq} baseline. BART has been shown to excel at semantic parsing with no need for special parametrization \cite{chen-etal-2020-low,bevilacqua2021one}. We train BART to produce the SPARQL directly from input sentences and we use the same training data as for TransQA. To incorporate Entity Linking, we concatenate BLINK entity linking to the input to teach BART to copy it into the target.

Table~\ref{tab:kbqa-results} displays the comparison with related work. The proposed TransQA outperforms all methods both on DBpedia and Wikidata, with two notable exceptions. EDGE-QA has better results in QALD-9 but it utilizes likely stronger entity linking. Also the simple BART baseline outperforms all other methods in simple web questions, while lagging clearly behind in LC-QuAD and QALD, which are more complex. This result may be explained by a trade-off between inductive bias and expressiveness, helping TransQA for complex grammars but limiting performance in simple cases. It is also worth noting that TransQA outperforms NSQA and SYGMA, while exploiting similar inductive bias and being simpler\footnotetext[5]{The relation linking method GenRel \cite{genrel} reports $59.6$ F1 result in LC-QuAD1.0 when combined with NSQA, without specifying the integration details.}\footnote[6]{STaG-QA SWQ-WD experiments use a self-selected data split, with smaller, non comparable test set}.

\section{Oracle and State Encoding Analysis}
\label{oracleperformance}

Table \ref{tab:oracle-results} reports oracle performance for BLINK and gold entity linking on the development set of LC-QuAD 1.0 and QALD and system performance of a TransQA trained with the corresponding oracle. Note that both results still use real AMR parses from APT~\cite{zhou-etal-2021-amr}. The gold entity linking oracle results measures how well oracle covers the distribution of SPARQL patterns found in these datasets and can be seen as an approximate upper bound for the performance of TransQA. Overal, correcting for entity linking errors, oracle performance reaches $78.7$ and $77.5$ for LC-QuAD 1.0 and QALD respectively. Although much higher than any reported system performance, this indicates oracles do not cover all possible SPARQL patterns, supporting the idea that a transpiling oracle implies a trade-off between inductive bias and expressiveness. Overall, this trade-off has strong positive effect if we compare the TransQA and BART seq2seq baseline results in Table~\ref{tab:oracle-results}.

Entity linking errors are shown to have a strong influence in TransQA, reducing performance by at least $5$ points for both LC-QuAD 1.0 and QALD. Other systems can also be expected to have a similar dependency since they can not recover for a linking error, making systems harder to compare. 

Table \ref{tab:ablation-results} ablates the inductive bias introduced in BART through attention and output vocabulary modifications described in Section~\ref{state}. Masking cross-attention to attend to the sentence words aligned to and AMR path, and thus to a potential triple, improves performance by around $4$ points for an optimal number of $8$ modified attention heads, selected on the dev set. Constrained decoding has an even larger effect of $10$ points on LC-QuAD 1.0 and $12$ points on QALD-9, sensibly larger than the effect of gold entity linking. This is likely because of the large reduction on possible decoding options. Another possibility is that it provides some visibility into the KG structure. This is particularly important to compensate for the fact that the knowledge graphs lack a schema, i.e. it is unclear if relations expressing one particular category will exist. Constrained decoding provides thus a limited view into the database, informing future decisions of the transpiler.

\section{Prior Work}

See Section~\ref{related}, for descriptions of NSQA \cite{kapanipathi-etal-2021-leveraging}, SYGMA \cite{neelam2021sygma}, EDGQA \cite{edgqa} and STaG-QA \cite{ravishankar2021twostage}. Both NSQA and SYGMA are relatively similar and involve a combination of multiple systems including entity and relation linking an integration module and a large set of rules. They can thus be seen as having the strongest inductive bias, exploiting human knowledge at the cost of complexity and limited expressiveness. The BART seq2seq baseline can be seen as the other extreme, lacking any inductive bias and clearly suffering when fine-tuned to complex grammars and limited amounts of train data. TransQA can be seen as finding the optimum between these two, needing only a small sub-set of NSQA's rules, and introducing the inductive bias through oracle imitation, attention masking and constrained decoding. The resulting system, is clearly simpler than NSQA and SYGMA but has higher performance in all corpora tested and, unlike BART, it does not suffer with more complex grammars. 

EDGQA and other recent approaches such as \cite{Saparina2021SPARQLingDQ}, transform natural language into intermediate question decomposition representations that are not well known and require of a number of rules. Here we employ AMR, a well established formalism with abundant training data in various languages and robust high performant parsers e.g. \cite{lee2021maximum}. TransQA is threfore likely to generalize better while requiring a small set of rules. Though EDGQA outperforms TransQA on QALD, it has to be taken into account that this model uses a ensemble of models for entity linking that likely provides extra performance.

Though less performant than TransQA, STAG-QA is also a relatively simple approach using two stages of inference. The latter step needs however the beam search which can be computationally prohibitive in datasets with more complex structure than LC-QUAD.

GGNN \cite{webqspwd} is a baseline system on WebQSP-WD dataset which specifically models the graph structure of the semantic parse using graph convolution. Is also defines a custom semantic parsing formalism to construct semantic graph representation from the train examples in WebQSP-WD. Our system outperforms GGNN by $7$ F1 points on the WebQSP-WD dataset.

More recently~\cite{banerjee2022modern}, propose a seq2seq baseline with T5 where they assume the gold entities and relations have been provided. As shown in our experiments in Table~\ref{tab:oracle-results}, gold entity linking alone causes a large difference in performance. This work is therefore not comparable with the rest of the approaches here considered.

\section{Conclusions}

We introduced TransQA, an approach for KBQA that transpiles a well established semantic representation, AMR, obtained from an off-the-shelf parser, into SPARQL. This transpiling operation is learned through a transition-based approach, resulting on a simpler and more performant system than prior approaches leveraging AMR. Empirical results on KBQA corpora show that TransQA finds an optimal trade-off between inductive bias and expressiveness, outperforming both sequence to sequence models and more complex pipeline systems incorporating a large number of rules. Ablation analysis shows that encoding of the transpiler state in the self-attention mechanism and constrained decoding yields clear gains, while removing the need to encode AMR. Finally oracle analysis reveals that the proposed transition system is able to attain high coverage across datasets while remaining simple.

\section{Limitations}
 
The current implementation has several limitation which can be addressed in future work. 
We show in section 5.1 that our technique is severely impacted by the quality of relation linking. This also complicates comparison across systems.
Furthermore, we still rely on a small set of rules, see Figure ~\ref{pathalgo} and do not support type constraints and other SPARQL constructs. This limits the coverage of the model, see Table~\ref{tab:ablation-results}, although overall still leads to competitive results and theoretical limits given by the oracle are close to $80$ F1. For Wikidata our system uses direct KB relations and the current implementation can not predict statements and reified relations.

Regarding data and compute requirements, the model is trained with small data-sets by leveraging transfer learning from BART and a pre-trained semantic parser. Both models have multiple other purposes, amortizing the overall cost.

The experiments were carried out for the English language but there are little language-specific choices. Both multi-lingual versions of BART \cite{liu2020multilingual} and cross-lingual transition AMR parsers \cite{lee2021maximum} exist and some of the corpora like QALD have multi-lingual versions. Though limited by the amount of data available and alignment quality, the approach could be in principle extended to other languages without drastic changes being needed.

\bibliography{anthology,custom}
\bibliographystyle{acl_natbib}




\end{document}